%% file: main.tex
\documentclass[10pt,twocolumn,letterpaper]{article}

\usepackage{cvpr}              

\input{preamble}

\definecolor{cvprblue}{rgb}{0.21,0.49,0.74}
\usepackage[pagebackref,breaklinks,colorlinks,citecolor=cvprblue]{hyperref}

\def\paperID{12} 
\def\confName{CV4AEC}
\def\confYear{2024}

\title{An Expeditious Spatial Mean Radiant Temperature Mapping Framework using Visual SLAM and Semantic Segmentation}

\author{
Wei Liang, Yiting Zhang, Ji Zhang, Erica Cochran Hameen\\
Carnegie Mellon University\\
5000 Forbes Ave, Pittsburgh, Pennsylvania, USA\\
{\tt\small \{weiliang, yitingzh, zhangji, ericac\}@andrew.cmu.edu}
}

\begin{document}
\maketitle
\input{sec/0_abstract}    
\input{sec/1_intro}
\input{sec/2_method}
\input{sec/3_results}
\input{sec/4_conclusion}
{
    \small
    \bibliographystyle{ieeenat_fullname}
    \bibliography{main}
}


\end{document}

%% file: preamble.tex
\usepackage[dvipsnames]{xcolor}
\usepackage{algorithm}
\usepackage{algpseudocode}


%% file: sec/0_abstract.tex
\begin{abstract}
Ensuring thermal comfort is essential for the well-being and productivity of individuals in built environments. Of the various thermal comfort indicators, the mean radiant temperature (MRT) is very challenging to measure. Most common measurement methodologies are time-consuming and not user-friendly. To address this issue, this paper proposes a novel MRT measurement framework that uses visual simultaneous localization and mapping (SLAM) and semantic segmentation techniques. The proposed approach follows the rule of thumb of the traditional MRT calculation method using surface temperature and view factors. However, it employs visual SLAM and creates a 3D thermal point cloud with enriched surface temperature information. The framework then implements Grounded SAM, a new object detection and segmentation tool to extract features with distinct temperature profiles on building surfaces. The detailed segmentation of thermal features not only reduces potential errors in the calculation of the MRT but also provides an efficient reconstruction of the spatial MRT distribution in the indoor environment. We also validate the calculation results with the reference measurement methodology. This data-driven framework offers faster and more efficient MRT measurements and spatial mapping than conventional methods. It can enable the direct engagement of researchers and practitioners in MRT measurements and contribute to research on thermal comfort and radiant cooling and heating systems.
\end{abstract}

%% file: sec/1_intro.tex
\section{Introduction}
\label{sec:intro}

Thermal comfort is defined as “\textit{the condition of mind which expresses satisfaction with the thermal environment and is assessed by subjective evaluation}”~\cite{ashrae2023standard}. The evaluation of thermal comfort involves physical measurements and subjective factors. Among these measurements, mean radiant temperature (MRT) represents the short- and long-wave radiant heat fluxes reaching the human body. MRT converts radiant heat exchange between the occupant and heterogeneous surrounding surfaces to a similar process between the human body and a uniform enclosure. Deviated MRT can cause thermal discomfort even if the local air temperature of the occupants is acceptable~\cite{DAmbrosio_Alfano2013-os}. 

Assessing MRT is challenging due to its geometric complexity~\cite{Kantor2011-iw} and the peculiarities of the instruments~\cite{DAmbrosio_Alfano2013-os}. Most MRT measuring methods, regulated in the ISO 7726 standard~\cite{ISO7726}, are time-consuming, labor-intensive, and require high expertise to handle instruments and perform calculations. Therefore, the MRT assessment is limited to individual measurements due to instrument and technique restrictions. As the \textbf{first contribution} of this paper, we proposed a novel data-driven MRT measurement approach using visual simultaneous localization and mapping (SLAM) and semantic segmentation techniques, enabling efficient MRT data collection. 

\begin{figure*}[t]
  \centering
   \includegraphics[width=0.8\linewidth]{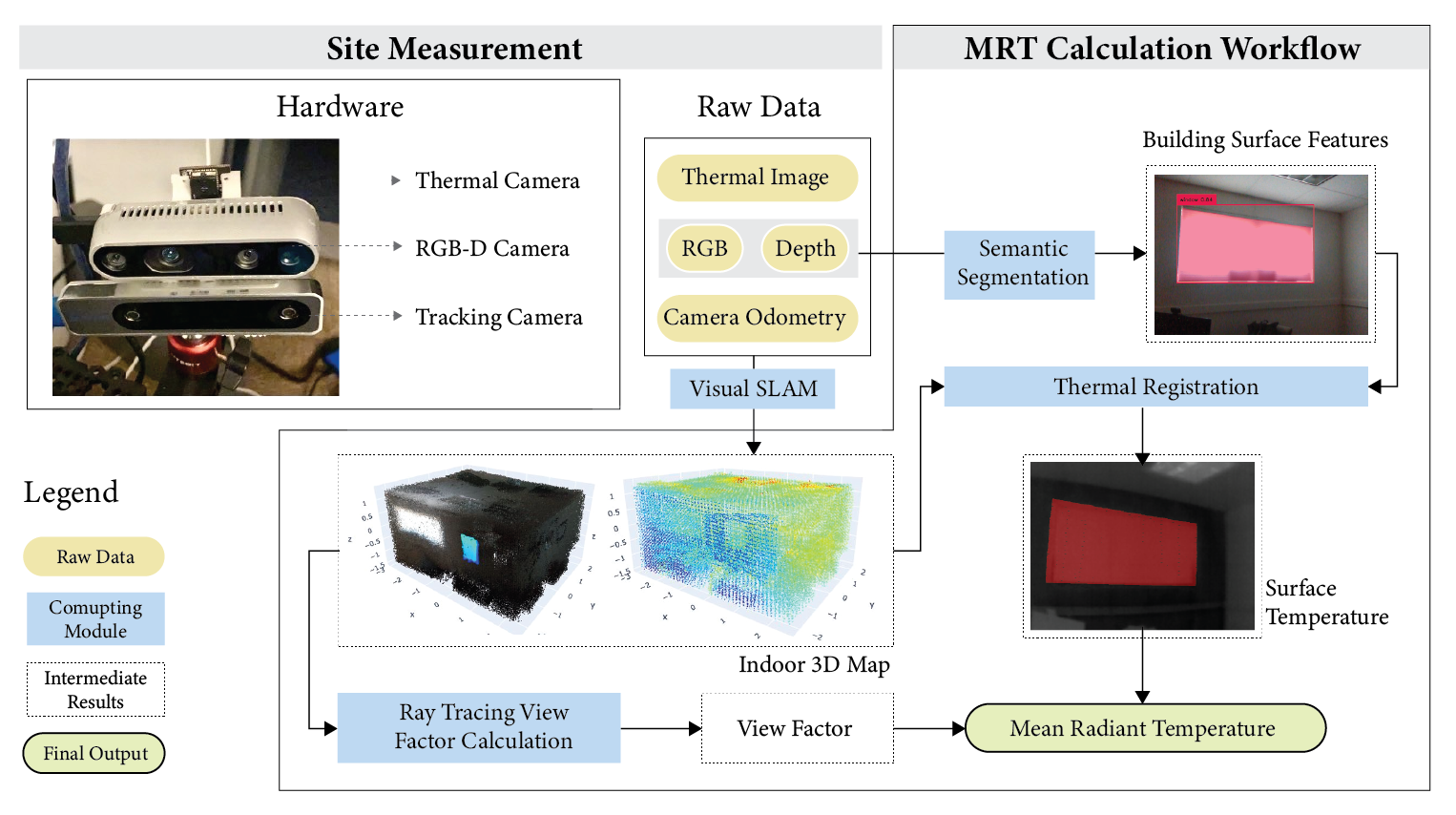}

   \caption{The block diagram of the proposed framework.}
   \label{fig:diag}
\end{figure*}

In addition, previous frameworks for spatial MRT rely on either sophisticated simulation software~\cite{aviv2019thermal,Hou2023-hf}, domain knowledge~\cite{teitelbaum2017mapping}, and advanced sensors~\cite{aviv2022data} to extract features. The \textbf{second contribution} of this paper is to provide an end-to-end framework using a low-cost camera array and semantic segmentations to extract features automatically. The block diagram of the proposed framework is shown in~\cref{fig:diag}. The GitHub repository will be published later.

%% file: sec/2_method.tex
\section{Method}
\label{sec:method}

\subsection{Camera Set Design and Thermal Calibration}

The camera array comprises a thermal infrared (TIR) camera, an RGB-D camera, and a tracking camera. The TIR camera used in this study is the FLIR Lepton 3.5~\cite{noauthor_lepton_nodate}. Previous distance calibration results\cite{Li2019-zs} have suggested that the FLIR Lepton 3.5 provides precise and consistent distance measurements within the range of 0.8 to two meters. The camera set also includes Intel\textsuperscript{\textregistered} RealSense\textsuperscript{\texttrademark} Tracking Camera T265 in tandem with the
Intel\textsuperscript{\textregistered} RealSense\textsuperscript{\texttrademark} Depth Camera D435i. All three cameras are rigidly coupled together, spatially aligned in a 3D-printed frame, and mounted on the top of a mobile sensing platform, as shown in~\cref{fig:diag}. 

\begin{figure}[t]
  \centering
   \includegraphics[width=0.85\linewidth]{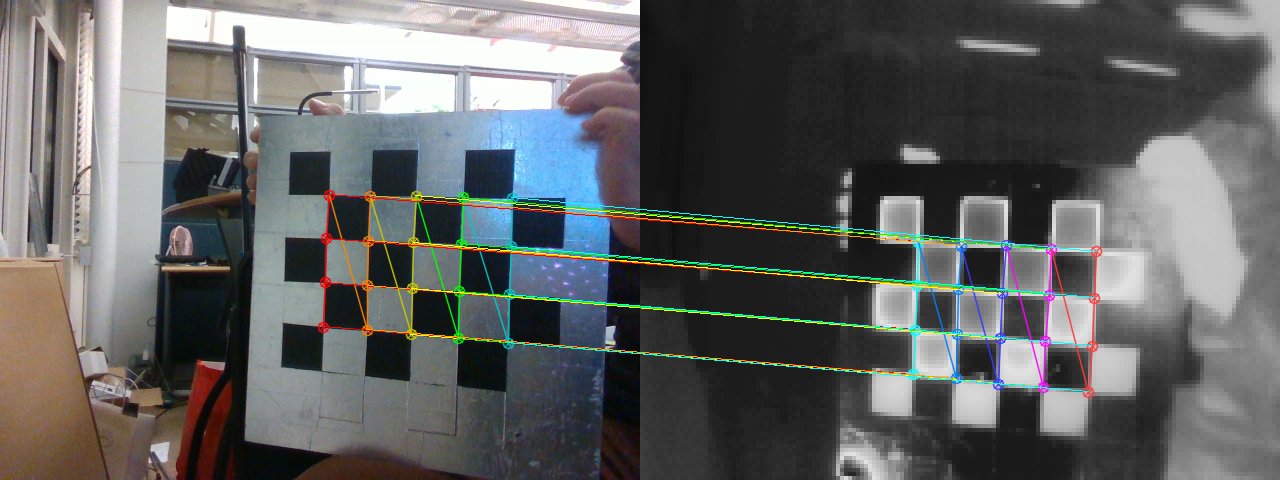}

   \caption{Left: RGB image. Right: thermal image.}
   \label{fig:cali}
\end{figure}

 The homogeneous transformation between the RGB-D camera and tracking can be found in its official guideline~\cite{schmidt2019intel}. However, stereo vision calibration between the RGB-D and TIR cameras is still needed for other modules in the proposed framework. Traditional calibration targets may not result in distinct geometric patterns visible in infrared images. Therefore, a specially designed checkerboard made of silver aluminum foil and black polyethylene material ensures conspicuous contrast in both visible and thermal ranges. As illustrated in \cref{fig:cali}, when heated by a radiator, the aluminum foil squares absorb additional infrared energy and appear dark, while the polyethylene squares emit more infrared energy and appear bright in the thermal image. The mean re-projection error (MRE) of 80 image pairs is 0.80 pixels. Similar TIR camera calibration approaches can be found in~\cite{liu2018thermal,Li2019-zs}.

\subsection{Visual SLAM and Thermal Data Fusion}

The proposed framework implements a visual simultaneous localization and mapping (vSLAM) system, enabling the camera array to incrementally build a map and localize itself in an unknown indoor environment. The vSLAM approach this framework applies is real‐time appearance‐based mapping (RTAB‐Map), given its loop closure detection feature and efficient memory management~\cite{Labbe2019-ly}.

Since the thermal infrared images are insufficient in frequency, field of view, and resolution, the thermal data fusion is implemented after the real-time measurement. At each synchronized timestamp $t$, the proposed framework maps thermal images with depth images and creates a local point cloud with temperature information. After that, the thermal point cloud is transformed to the world coordinates using the camera odometry at the current timestep. A K-D Tree-based iterative closest point algorithm~\cite{Rusinkiewicz2001-wt} aligns and merges local thermal point clouds to a 3D indoor thermal map. \cref{fig:pc} shows the RGB point cloud retrieved during the vSLAM process and the thermal point cloud retrieved offline.

\subsection{Semantic Segmentation for Thermal Features}
The primary reason for potential significant inaccuracies in MRT calculation using surface temperatures and view factors is that building envelopes in built environments have non-uniform surface temperatures. Walls, floors, ceilings, and other surfaces may contain different features, such as windows, paintings, lighting, skylights, and other building components, leading to different temperature profiles across different surface areas. Therefore, we must detect and segment the building surface features from the main surface areas. To achieve this goal, we implement Grounded SAM~\cite{Ren2024-nw}, which uses Grounding DINO~\cite{Liu2023-jn} as an open-set object detector to combine with the segment anything model (SAM)~\cite{Kirillov2023-zx}. It provides open-vocabulary object detection, enabling the recognition of various building surface features that are not prevalent in most common datasets. 

When vSLAM processes sequential frames to generate a point cloud, each instance segmentation result is assigned a label or multiple labels over time based on outputs from the object detector. These labels correspond to different building surface features identified in the point cloud. Each feature is uniquely identified, and then "votes" for its label are based on the frequency of each label it has been assigned. The label that receives the most votes is selected as the final label for that feature. This approach ensures consistency and stability in the segmentation results by favoring the most common labels over time.

\begin{figure}[t]
  \centering
   \includegraphics[width=0.85\linewidth]{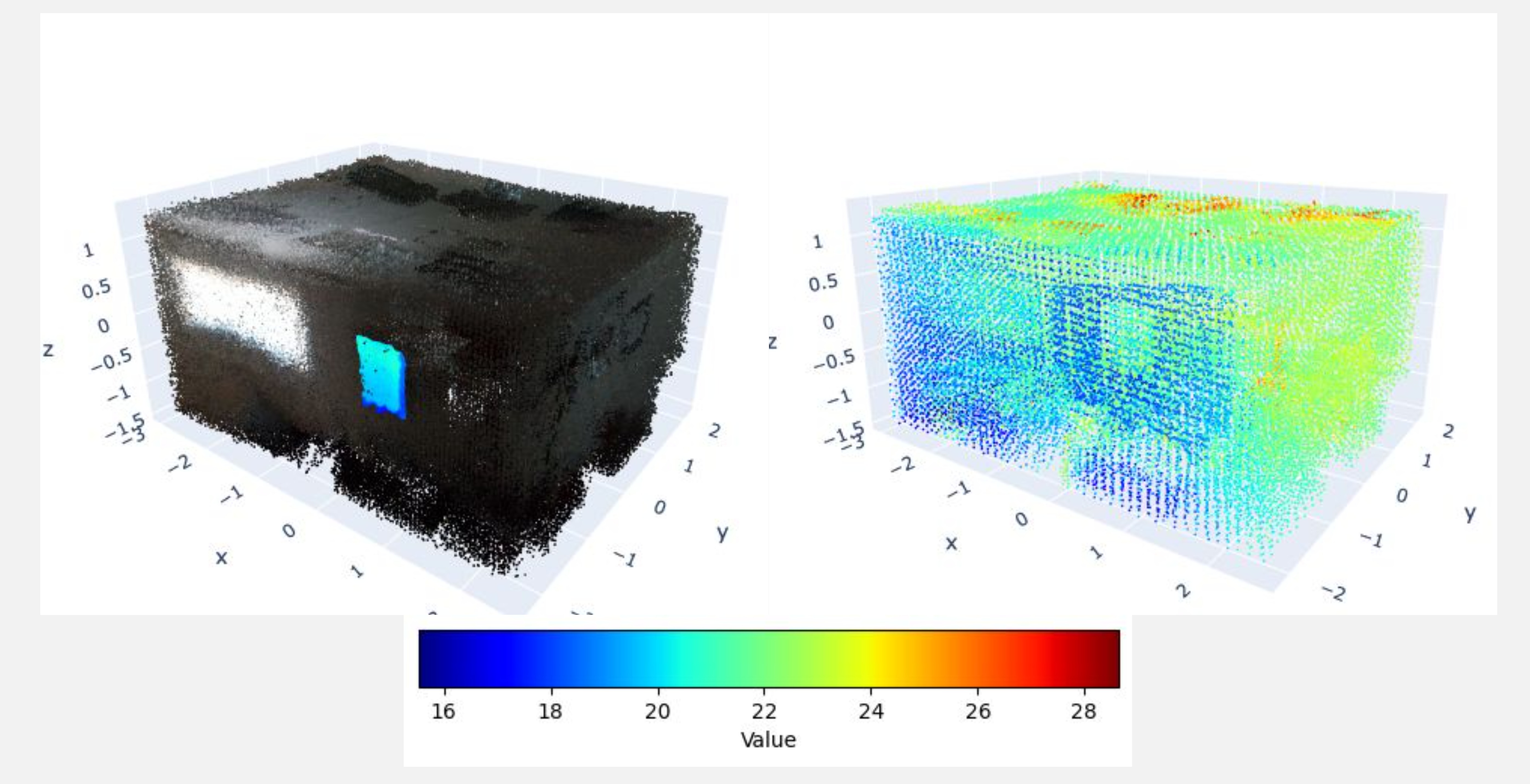}
   \caption{Left: RGB point cloud with a registered thermal feature, painting. Right: thermal point cloud.}
   \label{fig:pc}
\end{figure}

After segmentation, each feature's temperature profile is re-projected to the thermal point cloud. On the left of \cref{fig:pc}, the temperature profile of the feature "painting" on the wall is re-projected to the point cloud for view factor calculation.

\begin{figure}[t]
  \centering
   \includegraphics[width=0.85\linewidth]{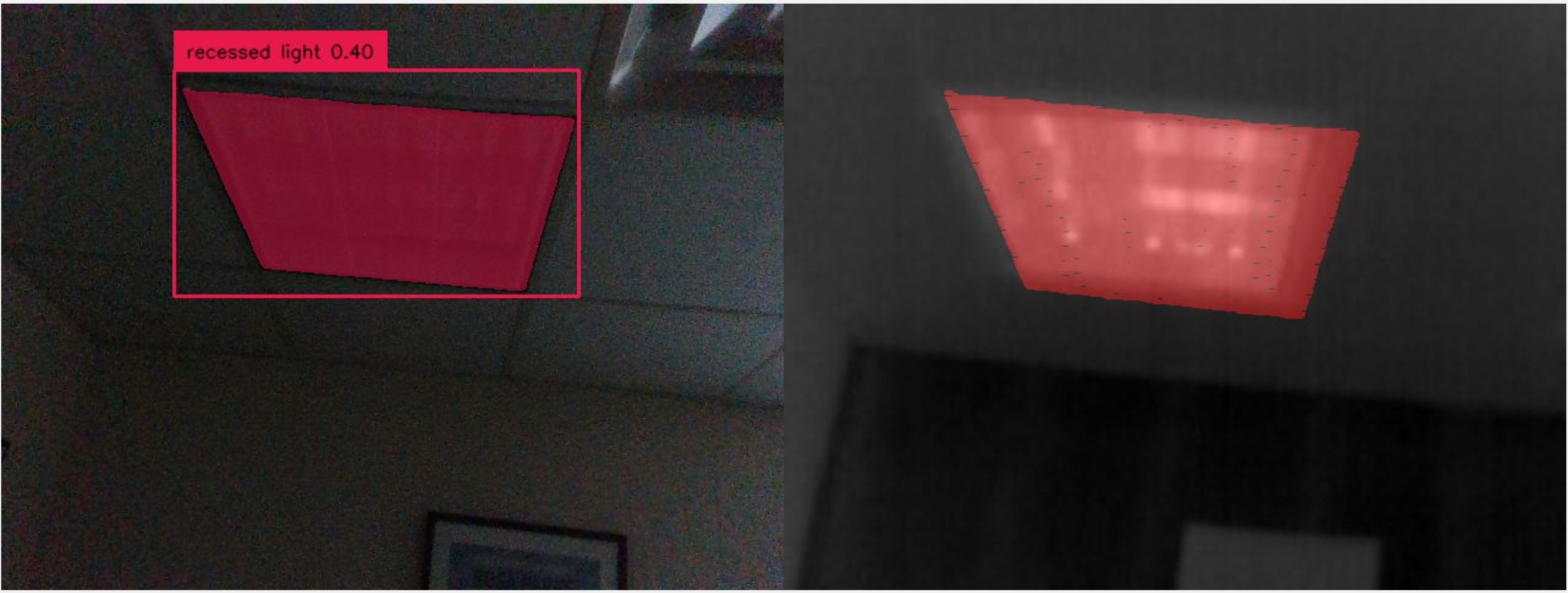}

   \caption{Detection and segmentation of ceiling feature "recessed lighting" on RGB image, and registration on the thermal image at approximately the same timestamp.}
   \label{fig:feature}
\end{figure}

An example of thermal feature segmentation is shown in~\cref{fig:feature}. We defined a label named "recessed lighting" to extract the related features on the ceiling. Grounded SAM detected and segmented this object type successfully. Using depth images that share the same frame as the RGB image, we can register the feature on the corresponding thermal image, where the recessed lighting exhibits a lighter color than its surrounding ceiling tile, indicating that it has a different temperature profile. 

\subsection{Mean Radiant Temperature Calculation}
The proposed framework for MRT measurement is based on the calculation approach from the temperature of the surrounding surfaces and the view factors between the observation point and the surrounding surfaces per ISO 7726~\cite{ISO7726}. The following \cref{eq:mrt} is used:

\begin{equation}
    \Bar{T}_{r} = \sqrt[4]{\frac{\sum_{i=1}^{n} F_i \cdot T_i^4}{\sum_{i=1}^{n} F_i}},
    \label{eq:mrt}
\end{equation}
Where $\Bar{T}_{r}$ is the mean radiant temperature in kelvins, $T_i$ is the surface temperature of surface or feature $i$ in kelvins, and $F_i$ is the view factor between the observation point and surface or feature $i$.

In the MRT calculation process, view factor calculation plays an indispensable role. We implement the Monte Carlo-based ray tracing algorithm to estimate view factors. A large number of rays are simulated, and the view factor is calculated by the proportion of rays intersecting with surfaces and features.

The mean radiant temperature calculation algorithm is provided in~\cref{alg:mrt}.

\begin{algorithm}
	\caption{Mean Radiant Temperature Calculation}\label{alg:mrt} 
	\begin{algorithmic}[1]
		\For {Each building envelope $i$}
                \State Retrieve surface plane and corners
                \State Calculate the initial view factor of the envelope $F_i$
                \If {Thermal features are found}
                \For {Each thermal feature $j$}
    				\State Retrieve surface corners
                        \State Calculate the view factor of the feature $F_{i,j}$ 
    				\State Compute feature surface temperature $T_{i,j}$
                    \State Correct the view factor of the envelope: $F_i =F_i -F_{i,j}$
			        \EndFor
                \EndIf
                \State Compute substracted building envelope surface temperature  $T_i$
            \EndFor
        \State Calculate MRT using~\cref{eq:mrt}
	\end{algorithmic} 
\end{algorithm}

%% file: sec/3_results.tex
\section{Results and Discussions}

The experiment was conducted on February 6th, 2024, in an enclosed office room in Pittsburgh, Pennsylvania, USA. The outdoor air temperature is about 1.1$^\circ C$. The indoor air setpoint is 21.0$^\circ C$. The wall facing east has a window, while the remaining three walls are connected to an open office space that shares the same HVAC system as the test environment. The handheld visual SLAM procedure took approximately 343 seconds to complete. Due to the short duration, we assume no temporal variations in MRT for this scenario.

\subsection{Spatial Field Mapping of MRT}
\begin{figure}[t]
  \centering
   \includegraphics[width=0.95\linewidth]{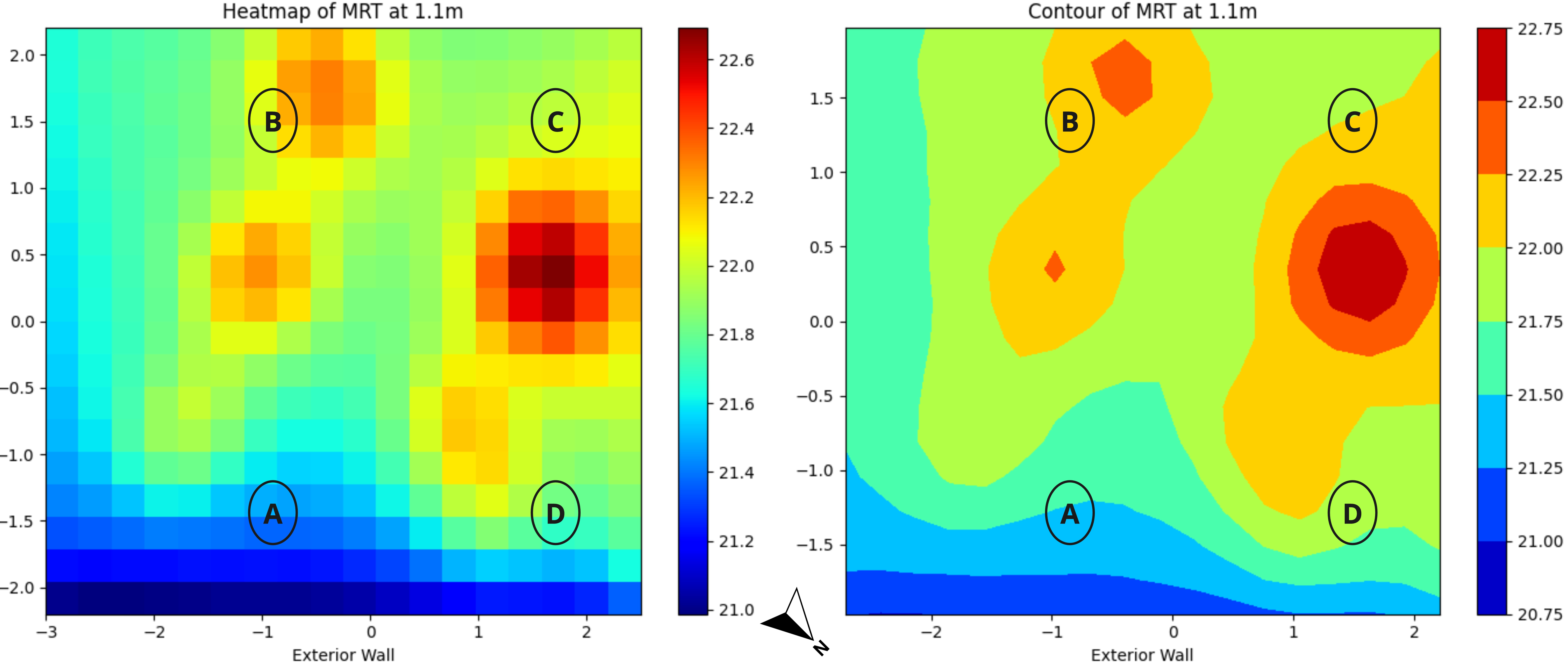}

   \caption{The heatmap and contour representations of the MRT radiant field were reconstructed using the proposed framework. Locations labeled A through D are the measurement points using a black-globe thermometer.}
   \label{fig:mrt_map}
\end{figure}

To generate the spatial field mapping of the MRT during the experiment, a 20$\times$20 mesh grid is applied to the indoor environment, and the MRT at each location is calculated using~\cref{eq:mrt}. The height of the MRT measurement is set at 1.1 meters above the floor, which is the height of the head and eyesight of occupants when sitting. As shown in~\cref{fig:mrt_map}, we can learn that the outdoor air temperature and the location of windows impact the spatial distribution of MRT. The area near the exterior wall exhibits lower MRT. However, a gradient ascent of MRT is in line with the increased directional solar exposure. We can also observe distinct warm zones near location C with a steep temperature gradient, indicating its proximity to heat-generating features, namely the recessed lights. In summary, the spatial distribution map of MRT provides more insights into the spatial variation of thermal comfort than individual measurements.

\subsection{Results Validation with Reference Method}
During the experiment, we also measured MRT using the black-globe thermometer as the reference measurement at four locations. The black-glob thermometer is considered a standard MRT measuring method, but it needs a long response time, and the instrument is bulky. The MRT measured from this reference method is:
\begin{equation}
    T_{mrt} = \sqrt[4]{T_{g}^4 + 0.4\times 10^8 (T_{g} - T_{a})\sqrt[4]{|T_{g} - T_{a}|}},
    \label{eq:globe}
\end{equation}
where $T_{g}$ is the globe temperature measured by the black-globe thermometer and $T_{a}$ is the local air temperature. The globe thermometer is placed at 1.1 meters above the floor. Note that each measurement takes 15 minutes for the globe thermometer to respond to the surroundings.

As shown in~\cref{tab:example}, the reference observations were conducted across four distinct locations, labeled A through D in~\cref{fig:mrt_map}. The data reveal a slight variance between the two sets of measurements, with the differences of MRT measurements in Celsius ($\Bar{t}_r$) ranging from -0.4 to +0.6$^\circ C$. Specifically, locations B and C exhibited higher MRT values as measured by our method compared to the reference, indicating a potential overestimation by our framework in these instances. The MRT measurement deviations at all locations comply with the range of the required accuracy ($\pm 2^\circ C$) specified by ISO 7726~\cite{ISO7726}, though not within the desired range ($\pm 0.2^\circ C$). Since each measurement of the reference method possesses a long response time, the temporal differences or methodological nuances may contribute to the observed variances.

\begin{table}
  \centering
  \begin{tabular}{@{}lccc@{}}
    \toprule
    Location & Black-globe & Ours & $\Delta \Bar{t}_r (^\circ C)$ \\
    \midrule
    A & 21.8$^\circ C$ & 21.4$^\circ C$  & -0.4 \\
    B & 21.6$^\circ C$ & 22.2$^\circ C$ & +0.6 \\
    C & 21.5$^\circ C$ & 22.0$^\circ C$ & +0.5 \\
    D & 21.1$^\circ C$ & 21.5$^\circ C$ & +0.4 \\
    \bottomrule
  \end{tabular}
  \caption{Comparison between the MRT measurements by the proposed framework and the reference method.}
  \label{tab:example}
\end{table}

\subsection{Limitations and Future Work}
Several limitations are involved in the proposed framework and experiment presented. Firstly, vSLAM can drift. Loop closure may not be guaranteed when the framework is applied to a larger indoor environment. Future improvements may involve using 3D-lidar and implementing pose graph optimization. Another limitation is that the experiment only compares the proposed framework with the black-globe method, which is coupled with a temporal discrepancy in the readings due to the long response time inherent to the black-globe method. Future work will broaden the comparative analysis with more standard MRT measurement methodologies, including the contact thermometer measurement approach and the calculation methods based on surface areas.

%% file: sec/4_conclusion.tex
\section{Conclusion}
This paper proposed an efficient data-driven mean radiant temperature measuring method using an RGB-D-TIR-Tracking camera array, visual SLAM, and semantic segmentation. The proposed framework detects and segments distinct building envelope surfaces and features in the built environment. It then calculates the corresponding view factors from the point of view using the Monte Carlo-based ray tracing method. The framework can calculate the MRT at any location in the indoor environment and generate a spatial distribution of MRT in an indoor environment based on detailed surface temperature information from the space. The proposed method can be used for research on thermal comfort and radiant systems and serve as part of the digital twin of a built environment for monitoring indoor environmental quality. 